\documentclass[letterpaper, 10 pt, conference]{ieeeconf}  

\IEEEoverridecommandlockouts                              

\usepackage[utf8]{inputenc}
\let\labelindent\relax
\usepackage{amsfonts}       

\usepackage{amsthm}
\usepackage{mathtools}      
\usepackage{amssymb}        
\usepackage{filecontents}
\usepackage{graphicx}       
\usepackage{marginnote}     
\usepackage{marvosym}       
\usepackage{overpic}        
\usepackage{tabularx}
\usepackage{subcaption}
\usepackage{cite}
\usepackage{color}
\usepackage{capt-of}
\usepackage{booktabs}
\usepackage[T1]{fontenc}
\usepackage[normalem]{ulem}
\usepackage{epstopdf}
\usepackage{enumitem}
\usepackage[normalem]{ulem}
\usepackage{lipsum}
\usepackage{url}
\usepackage[hidelinks]{hyperref}
\usepackage[dvipsnames]{xcolor}
\usepackage{diagbox}
\usepackage{multirow}
\usepackage{microtype}
\usepackage{xspace}
\usepackage{caption}
\makeatletter
\newif\if@restonecol
\makeatother

\usepackage[linesnumbered,ruled,vlined]{algorithm2e}
\usepackage{algpseudocode}
\usepackage{amsmath}

\newif\ifdraft
\draftfalse

\newif\ifarxiv
\arxivtrue
\arxivfalse

\ifdraft
\usepackage[paperheight=11in,paperwidth=9.5in,
			left=1.25in,right=1.25in,
			top=0.75in,bottom=0.75in,
			heightrounded,marginparwidth=1.2in,
			marginparsep=0.05in]{geometry}
\usepackage{xcolor}
\usepackage{xargs} 
\usepackage[textsize=footnotesize]{todonotes}
\newcommandx{\sh}[2][1=]{\todo[linecolor=blue,
			backgroundcolor=blue!10,bordercolor=blue,#1]{Han: #2}}
\newcommandx{\tg}[2][1=]{\todo[linecolor=orange,
			backgroundcolor=orange!10,bordercolor=orange,#1]{Greaten: #2}}
\newcommandx{\jy}[2][1=]{\todo[linecolor=green,
			backgroundcolor=green!10,bordercolor=green,#1]{JJ: #2}}
\else
\newcommand{\sh}[1]{{}}
\newcommand{\tg}[1]{{}}
\newcommand{\jy}[1]{{}}
\fi

\newif\iftwocolumn
\twocolumntrue

\def\ours{\textsc{RGBTrack}\xspace}

\makeatletter
\def\subsubsection{\@startsection{subsubsection}
                                 {3}
                                 {\z@ \hspace*{1mm}}
                                 {0ex plus 0.1ex minus 0.1ex}
                                 {0ex}
                                 {\normalfont\normalsize\itshape}}
\makeatother

\title{\LARGE \bf
RGBTrack: Fast, Robust Depth-Free 6D Pose Estimation and Tracking
}

\author{Teng Guo   \qquad Jingjin Yu
\thanks{G. Teng, and J. Yu are with the Department of 
Computer Science, Rutgers, the State University of New Jersey, Piscataway, NJ, USA. 
Emails: {\tt\small \{teng.guo, jingjin.yu\}@rutgers.edu}. This work was supported in part by NSF awards IIS-1845888, IIS-2132972, and CCF-2309866.
}
}

\begin{document}

\maketitle
\thispagestyle{empty}
\pagestyle{empty}

\begin{abstract}
We introduce a robust framework, \ours, for real-time 6D pose estimation and tracking that operates solely on RGB data, thereby eliminating the need for depth input for such dynamic and precise object pose tracking tasks. Building on the FoundationPose architecture, we devise a novel binary search strategy combined with a render‐and‐compare mechanism to efficiently infer depth and generate robust pose hypotheses from true-scale CAD models. To maintain stable tracking in dynamic scenarios, including rapid movements and occlusions, \ours integrates state-of-the-art 2D object tracking (XMem) with a Kalman filter and a state machine for proactive object pose recovery. In addition, \ours's scale recovery module dynamically adapts CAD models of unknown scale using an initial depth estimate, enabling seamless integration with modern generative reconstruction techniques. Extensive evaluations on benchmark datasets demonstrate that \ours's novel depth-free approach achieves competitive accuracy and real-time performance, making it a promising practical solution candidate for applications areas including robotics, augmented reality, computer vision. 

The source code for our implementation will be made publicly available at {\color{blue}{\url{https://github.com/GreatenAnoymous/RGBTrack.git}}}. 
\end{abstract}

\section{Introduction}
Determining and tracking in real-time an object’s 6D pose (i.e., 3D positions and orientations) from image sensor input is a cornerstone problem in spatial AI, with numerous applications including robotics, augmented reality, and many others. In robotics, for example, accurate pose information enables a robot to interact reliably with objects, facilitating fully automated processes such as assembly, warehouse pick-n-place operations, and so on.
Recent vision foundation models have demonstrated remarkable performance in pose estimation and real-time tracking, with FoundationPose~\cite{wen2024foundationpose} standing out as a prominant example. However, despite its strengths, FoundationPose depends on depth data for both registration and tracking. This reliance can lead to tracking loss when objects move rapidly or become occluded, and it produces inaccurate predictions when the CAD model is not at the true scale. As a result, integrating FoundationPose with modern generative object reconstruction methods~\cite{liu2024one, liu2023zero, liu2023one2345, jun2023shap}, which often yield CAD models with unknown scales, remains a significant challenge.

In this study, we address the challenge of model-based \emph{robust}, \emph{depth-free} 6D pose estimation which applies to objects with known CAD models, as well as to novel, unseen objects. 
Robust depth-free 6D pose estimation and tracking is advantageous in many practical scenarios. For example, in mobile augmented reality (AR) applications, where smartphones and wearables typically lack dedicated depth sensors, RGB-only methods enable immersive experiences without additional hardware. In outdoor robotics, RGB-based approaches can outperform depth sensors that struggle with sunlight and reflective surfaces. Moreover, for cost- and power-sensitive systems like small robots, drones, or embedded devices, relying on inexpensive RGB cameras simplifies design and reduces energy consumption. This approach also benefits legacy systems that already use standard cameras, avoiding the need for costly sensor upgrades. Finally, in cluttered or occluded environments, where depth data may be noisy or incomplete, advanced RGB-based techniques can provide more reliable pose estimation.

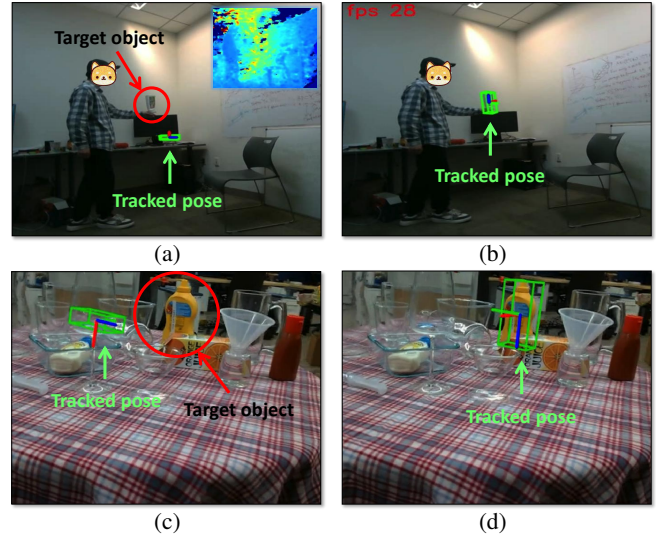
\begin{figure}[!t]
    \centering
    \begin{overpic}               
        [width=1\linewidth]{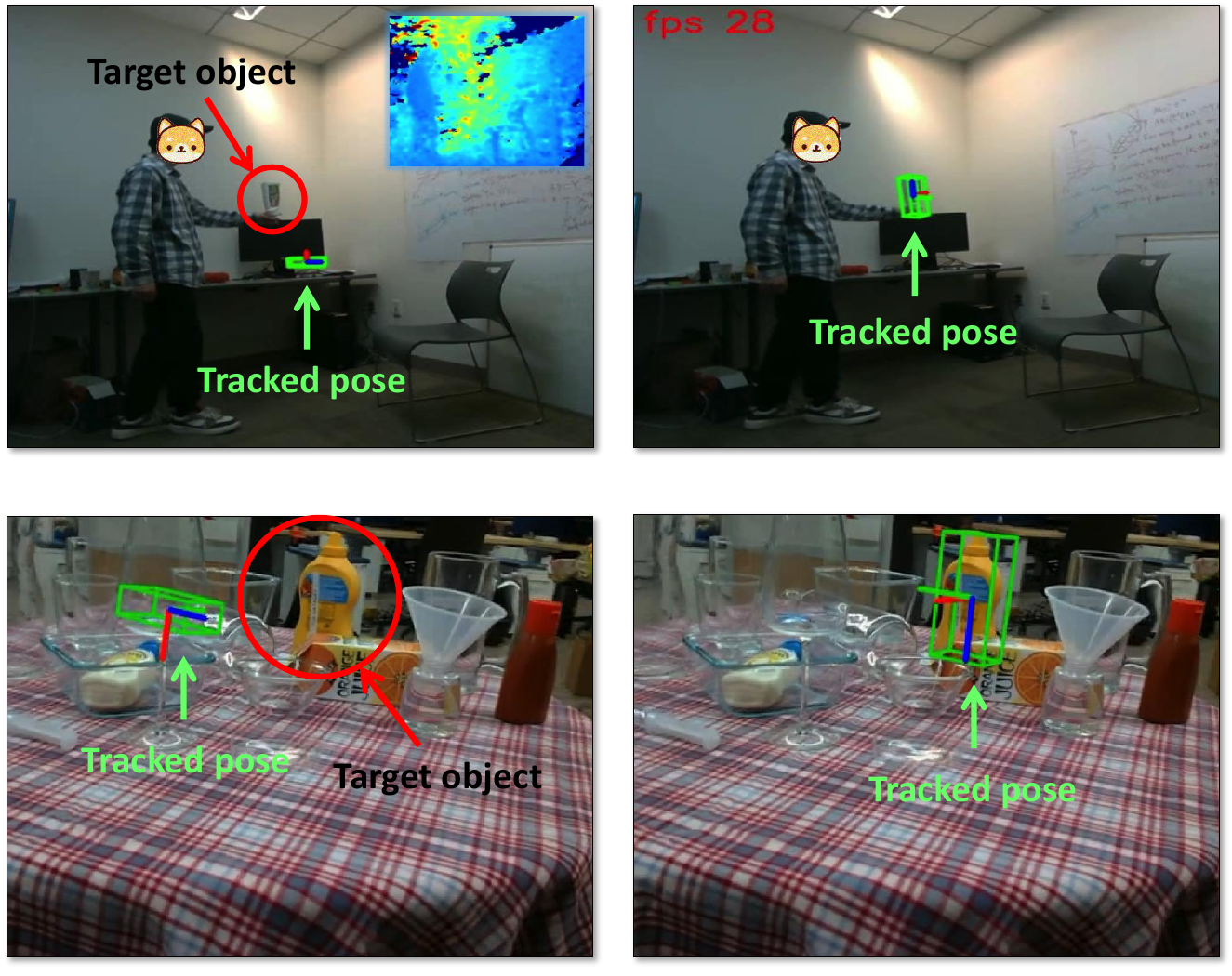}
             \small
             \put(22.5, 39) {(a)}
             \put(72.5, 39) {(b)}
             \put(22.5, -2.3) {(c)}
             \put(72.5, -2.3) {(d)}
        \end{overpic}
        \vspace{-2mm}
    \caption{Illustration of advantages of \ours. (a) When the depth input is of low quality, e.g., when objects are far away, methods such as FoundationPose~\cite{wen2024foundationpose} can struggle to generate and maintain accurate pose estimations. (b) Because RGB sensors have much higher resolution, \ours can track objects effectively, even without reliable depth data.
(c) State-of-the-art methods~\cite{wen2024foundationpose}, after losing  track of object temporarily due to occlusion of fast movement, struggle to correctly recover the tracked pose.
(d) \ours can effectively keep track of objects in the event of due to occlusion or rapid motion. }
    \label{fig:example}
         \vspace{-6mm}
\end{figure}

Indeed, our proposed method,\ours, excel in the above-mentioned settings (see, e.g., Fig.~\ref{fig:example} for some examples). Specifically, \ours enables:

\begin{itemize}
\item \textbf{Real-Time Depth-Free Pose Estimation and Tracking}: \ours operates \emph{exclusively on monocular RGB images}, using true-to-scale CAD models to generate a broad array of pose hypotheses to match the observed pose. The strategy not only improves computational efficiency but also maintains high precision in 6D pose estimation.
\item \textbf{Robust Tracking and Recovery}: Through the organic integration of FoundationPose and cutting-edge 2D object tracking networks like XMem~\cite{cheng2022xmem}, and the incorporation of a Kalman filter, \ours significantly \emph{boosts 6D object pose tracking stability}. A state machine is further employed to promptly detect object tracking loss which subsequently activates a recovery mechanism, enabling the system to reinitialize tracking even under the most challenging conditions.
\item \textbf{Handling CAD Models of Unknown Scale}: \ours encapsulates support for \emph{CAD models with unknown scale}, an essential capability in real-world applications where precise object dimensions are often unavailable. By utilizing an initial depth measurement from the first frame, \ours dynamically scales the CAD model to align with the input depth information, ensuring accurate pose estimation even with imprecise models.
\end{itemize}

Overall,  \ours delivers robust, real-time, depth-free 6D pose estimation and tracking without the need for extensive retraining, making it ideally suited for real-world deployment. Experiments demonstrate that the method consistently maintains high tracking accuracy even in challenging scenarios—such as under occlusions, during rapid movements, and when handling objects at multiple scales.

\begin{figure*}
    \centering
    \begin{overpic}[width=1\linewidth]{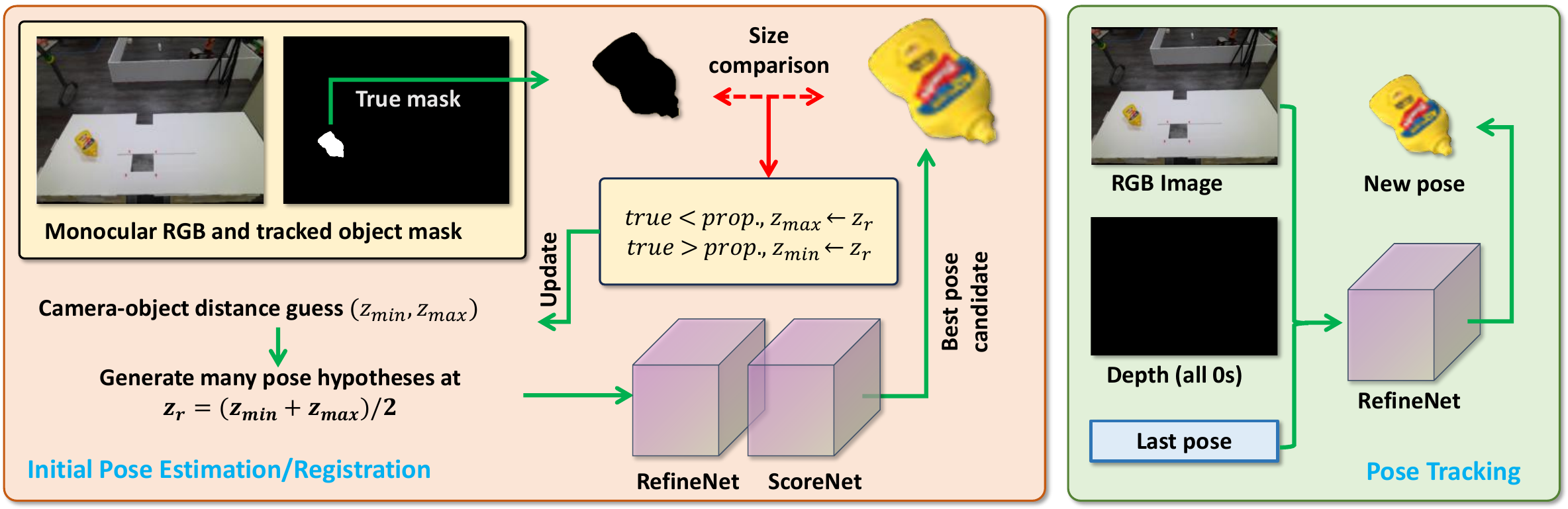}
        \small
    \end{overpic}
    \caption{Sketch of the \ours framework. [left] To perform initial pose estimation/registration without depth, object CAD is used to render many pose hypotheses assuming a camera-object distance of $z_r = (z_{\min} + z_{\max})/2$. These pose hypotheses are then refined and filtered using RefineNet and ScoreNet~\cite{wen2024foundationpose} to yield a single best pose candidate. Comparing the object mask at the candidate pose with the true object mask then allows a binary-search like procedure to obtain a pose that best matches the actual object pose in the RGB image. [right] Once an initial 6D pose estimate is obtained, RGB features and the previous pose estimate are sufficient for maintaining good pose tracking quality, using fake depth input of all zeros.}
    \label{fig:binary_search}
\end{figure*}

\section{Related Work}
6D pose estimation, which involves predicting the 3D rotation and 3D translation of an object relative to a camera, has been extensively studied in computer vision. Recent methods can be broadly categorized into feature matching-based and template matching-based approaches.
Feature-matching methods~\cite{peng2019pvnet,hu2020single,hu2019segmentation,li2019cdpn,rad2017bb8,tekin2018real, zakharov2019dpod, wen2020se, wen2021bundletrack, sun2022onepose, he2022onepose++} extract local features from an image, establish correspondences with a given 3D model, and estimate the 6D pose using a variant of the PnP algorithm \cite{moreno2007accurate} to recover the 3D-to-2D transformation. These approaches leverage local or global descriptors to match features across modalities.
Classical methods in this category include the \textit{Iterative Closest Point (ICP)} algorithm \cite{besl1992method}, which iteratively aligns a 3D model to an observed point cloud by minimizing point-to-point distances. Another widely used approach is the \textit{Point Pair Feature (PPF)} framework \cite{drost2010model}, which encodes geometric relationships between point pairs to facilitate pose estimation.
More recent learning-based methods, such as \textit{PVNet} \cite{peng2019pvnet}, employ deep networks to predict keypoint correspondences, making them robust to occlusions and clutter. Other hybrid approaches like \textit{CDPN} \cite{li2020cdpn} combine feature-based learning with direct pose regression to further improve accuracy.
Template matching-based approaches~\cite{ausserlechner2024zs6d, wen2024foundationpose, labbe2022megapose,cai2022ove6d, lin2024sam, nguyen2022templates, nguyen2024gigapose, okorn2021zephyr, ornek2024foundpose} directly compare observed images with a set of precomputed templates or renderings to determine the best-matching pose. \textit{PoseCNN} \cite{xiang2017posecnn} is an early deep learning-based method that regresses object poses from RGB images by treating pose estimation as a direct mapping problem. 
Foundation models based on template matching, such as \textit{FoundationPose}, refine initial pose estimates by iteratively aligning a rendered 3D model with the observed image through a learned optimization process.
%
Both categories of methods have demonstrated effectiveness in different scenarios. Feature matching-based methods are often more robust to occlusions. Still, they may struggle in textureless environments, while template matching-based approaches can leverage global scene information but require extensive training data and computational resources.

\textbf{Organization.} 
This paper is organized as follows. In Sec.~\ref{sec:prelim}, we review the basic framework of FoundationPose. In Sec.~\ref{sec:no_depth}, we introduce our novel depth-free pose estimation and tracking method. In Sec.~\ref{sec:robust_tracking}, we introduce our tracking loss recovery mechanism. Sec.~\ref{sec:incorrect_scale} briefly explains how to deal with improper CAD scales. In Sec.~\ref{sec:eval}, we perform evaluations on the YCBinEAOT dataset~\cite{wen2020se} and ClearPose dataset~\cite{chen2022clearpose}.  We conclude in ~\ref{sec:conclusion}.
%

\section{Preliminaries}\label{sec:prelim}
As \ours builds upon FoundationPose~\cite{wen2024foundationpose}, we provide a brief overview of its architecture. FoundationPose operates as a unified framework for 6-DoF object pose estimation and tracking, addressing both model-based and model-free setups through the following components:
\begin{itemize}
    \item \textbf{Neural Implicit Representation:} 
    In the model-free setup, where a CAD model is unavailable, FoundationPose employs an object-centric neural field representation. This representation enables high-quality novel view synthesis by encoding the object’s geometry as a signed distance field (SDF)~\cite{wen2023bundlesdf} and the appearance through view-dependent color rendering. For model-based setups, CAD models are directly utilized. In this work, we focus on the scenario when the CAD models are available.

    \item \textbf{Pose Initialization:}
    Given an input RGB-D image, a pre-trained detection network identifies the object's bounding box. The object’s translation is initialized based on the median depth within the bounding box, and rotation hypotheses are generated by uniformly sampling orientations around the object.

    \item \textbf{Pose Refinement:}
    Coarse pose estimates are refined iteratively using a neural network that compares renderings of the object (conditioned on the estimated pose) with the observed input. This process minimizes the misalignment by optimizing both translation and rotation updates in the camera frame.

    \item \textbf{Pose Selection:}
    A hierarchical pose ranking network evaluates multiple refined hypotheses. The module uses contrastive learning with a two-level comparison strategy: first between individual hypotheses and the observation, and then among all hypotheses collectively. The hypothesis with the highest ranking is selected as the final pose.
\end{itemize}

The unified framework allows for efficient pose estimation and tracking, achieving state-of-the-art performance in model-based and model-free scenarios across public benchmarks. 

\section{Depth-Free Pose Estimation and Tracking}\label{sec:no_depth}
Similar to FoundationPose, to enable depth-free pose tracking, \ours must obtain an initial pose estimation of the target object. Then, pose updates are made to keep track of the target object. We now outline how \ours executes these two steps without depth sensor input. The overall \ours pose estimation and tracking framework is depicted in Fig.~\ref{fig:binary_search}.
\subsection{Pose Estimation Without Depth Sensor Data}
Initial pose estimation is also known as pose registration. The \ours pose estimation/registration process is outlined in Fig.~\ref{fig:binary_search} on the left. In the absence of depth sensor input, estimating and tracking the 6-DoF pose of an object requires innovative approaches to infer depth. We achieve this goal by augmenting pose hypothesis generation to account for missing depth information.
In FoundationPose, rotations are sampled uniformly in the SO(3) space, while the object’s 3D position \((x, y, z)\) is estimated using the center of the mask, the median depth within the mask, and the camera intrinsics. The depth component \(z\) is computed as:

\begin{equation}
\label{eq:z_rough}
    z_c = \text{median}(\text{depth}[\text{mask}]),
\end{equation}
and the \(x\) and \(y\) coordinates are obtained using:

\begin{equation}\label{eq:xy_rough}
    x_c = \frac{z(u_c - c_x)}{f_x} \quad y = \frac{z(v_c - c_y)}{f_y},
\end{equation}
where \((u_c, v_c)\) denotes the pixel coordinates of the mask center, \((c_x, c_y)\) are the principal point coordinates, and \(f\) is the focal length.

When depth input is unavailable, a rough depth estimate must be obtained first. While recent metric depth estimation models such as ZoeDepth~\cite{bhat2023zoedepth} and Metric3D~\cite{yin2023metric3d} can provide depth predictions, their scale and shift inaccuracies often make them unreliable for direct use. An alternative approach is to sample pose hypotheses for both rotation and depth. If the object's depth lies within a known range \([\texttt{z\_min}, \texttt{z\_max}]\), we can sample \(N_z\) depth values uniformly within this range. For each sampled depth, we generate \(N_r\) rotational hypotheses, leading to a total of \(N_z N_r\) pose hypotheses. These are then refined and evaluated using the original pose selection network. However, this brute-force approach significantly increases computational demand, leading to potential CUDA memory shortages.

To overcome these limitations, we propose a more efficient \emph{binary search} algorithm that iteratively refines the depth estimate by enforcing consistency between rendered and observed image features.

The proposed algorithm estimates the pose of an object using binary search to determine the object's depth in the absence of direct depth measurements. Given an estimated initial pose, a 3D object mesh, the first frame RGB image, the initial object mask, and the camera intrinsic matrix $K$, the algorithm iteratively refines the pose estimate.
To start, the algorithm initializes the depth search range with predefined minimum and maximum depth values. At each iteration, it computes a midpoint depth $z_r$ and projects the object's center pixel coordinates onto the 3D space using the camera intrinsics. The algorithm then samples multiple candidate rotations and combines them with the estimated translation to form a set of candidate poses.

To refine the poses, we employ the \texttt{RefineNet} of FoundationPose, which optimizes the sampled poses, followed by running the \texttt{ScoreNet}, which selects the best pose among the refined candidates. The best pose is then used to render a synthetic mask of the object. The algorithm updates the depth estimate based on the difference between the rendered mask and the observed mask. If the rendered mask's area is larger than the observed mask, the search range is adjusted by setting the lower bound to the current midpoint; otherwise, the upper bound is updated.

The process continues until the estimated depth converges (i.e., the change in depth between iterations is below a threshold) or the search range becomes sufficiently small. The final pose estimate is returned once the stopping criteria are met.

\begin{algorithm}[!htbp]
\caption{Binary Search for Pose Registration Without Depth Sensor Data }
\label{alg:binary_search}
\KwIn{$ mesh model, rgb image, mask, K$}
\KwOut{$pose$}
\KwData{Initialize $low \gets \texttt{z\_min}$, $high \gets \texttt{z\_max}$, $last\_depth \gets \infty$}
\While{$low \leq high$}{
    $z_r \gets \frac{low + high}{2}$\;
    $p_x, p_y \gets$ center pixel of the mask\;
    $x_r, y_r \gets z_r (p_x - c_x)/f, z_r (p_y - c_y)/f$\;
    $poses \gets \texttt{SampleRotationVectors()} \oplus [x_r, y_r, z_r]$\;
    $poses \gets \texttt{RefineNet}(poses)$\;
    $pose \gets \texttt{ScoreNet}(poses)$\;
    $mask_r \gets \texttt{render\_cad\_mask}(pose, mesh, K, w, h)$\;
    $current\_depth \gets pose[2, 3]$\;
    
    \If{$\lvert current\_depth - last\_depth \rvert < 10^{-2}$}{
        \textbf{break}\;
    }
    $last\_depth \gets current\_depth$\;
    
    \If{$\lvert high - low \rvert < 10^{-2}$}{
        \textbf{break}\;
    }
    $area \gets \texttt{sum}(mask_r)$\;
    \If{$area > \texttt{sum}(mask)$}{
        $low \gets mid$\;
    }
    \ElseIf{$area < \texttt{sum}(mask)$}{
        $high \gets mid$\;
    }
}
\Return{$pose$}
\end{algorithm}

\subsection{Pose Tracking without Depth Sensor Data}
Assuming pose registration for the first frame is sufficiently accurate, we propose a simple method for tracking objects without relying on depth information. The \ours pose tracking procedure is outlined in Fig.~\ref{fig:binary_search} on the right. 

Because pose changes between frames are likely small, one might be attempted to use depth from the previous frame (rendered using the CAD model and estimated pose) as the input depth for the current frame. However, experimenting with the idea shows that it introduces significant drift, ultimately degrading tracking performance.
Other methods, such as predicting depth based on object movement using, e.g., a Kalman filter, also suffered from pose drifting issues.

As it turns out, instead, we can use a zero-depth matrix as input to FoundationPose to effectively mitigate pose drifts while maintaining good tracking performance.
The reason for this is interesting: FoundationPose’s RefineNet processes a six-channel input, where the first three channels are RGB, and the last three represent the point cloud, which depends on depth. The encoder extracts features from both RGB and point cloud data, combining them through a weighted sum. Depth features enhance the RGB representation, but when depth is zero, the model switches to rely entirely on the RGB features to compute the relative pose between two images. 

\begin{figure}[t!]
    \centering
    \begin{overpic}               
        [width=1\linewidth]{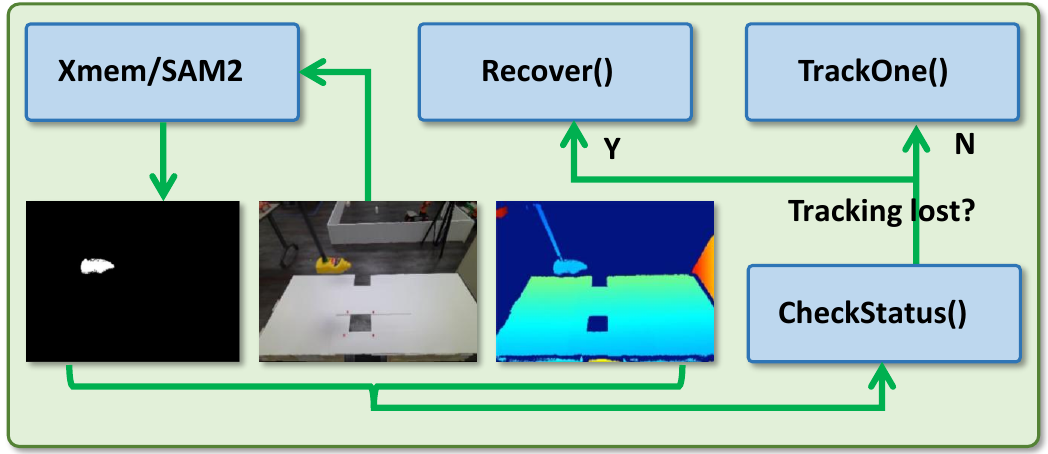}
             \small
        \end{overpic}
    \caption{The state-machine for pose tracking loss recovery. During each pose update, the mask of the tracked object is obtained, which is used to monitor object tracking status and detect potential tracking loss. If no loss is detected, then it is business as usual; otherwise, a pose recovery procedure is executed to maintain a ``fake'' pose while efforts are attempted to recover the pose. }
    \label{fig:recover_framework}
\end{figure}

\section{Real-Time Robust Pose Tracking with Recovery}\label{sec:robust_tracking}
To illustrate how \ours keeps track of object's 6D pose in the presence of occlusion and rapid pose changes, we first explain how this can be achieved using depth information. Then, we elaborate how the same can be achieved without using depth data.  
\subsection{Tracking Recovery with Depth}
To facilitate pose recovery in the event of pose tracking loss, we design a novel state machine (see Fig.~\ref{fig:recover_framework}) to manage object tracking status by integrating XMem with a Kalman filter. The system detects tracking failures and refines object poses using multiple prediction and correction strategies.

The process begins by verifying whether tracking is lost using the mask returned by XMem (or another 2D object tracking algorithm, such as SAM2~\cite{ravi2024sam}). If tracking is determined to be lost, the Kalman filter takes over to predict the object's position and orientation, updating and returning the estimated transformation matrix. Tracking is considered lost if either of the following two conditions is met:
\begin{itemize}
    \item The totality of the object mask, i.e., the total number of pixels inside the mask, falls below a specified threshold.
    \item Let $(x_c,y_c,z_c)$ be the coarse translation estimate from Eqs.~\eqref{eq:z_rough}-\eqref{eq:xy_rough}, and $(x_p,y_p,z_p)$ be the predicted pose by the baseline (FoundationPose). If the Euclidean distance between these estimates exceeds a threshold $\theta$, tracking is deemed lost:
    
    \begin{equation}\label{eq:lost_tracking_condition}
         \sqrt{(x_c-x_p)^2+(y_c-y_p)^2+(z_c-z_p)^2} > \theta.
    \end{equation}
       
\end{itemize}

If tracking is unreliable, the Kalman filter is updated, and multiple pose hypotheses are generated by sampling random poses. The object's center is estimated using depth and mask information. To maintain real-time performance, only 20 rotation hypotheses (as opposed to using 252 hypotheses in the initial pose registration phase) are sampled near the Kalman filter’s rotation prediction, followed by pose refinement using the refiner. A scoring network ranks the hypotheses and selects the best pose. If the new pose significantly deviates from the previous estimate, the tracking state is adjusted accordingly.

If none of the tracking failure conditions are met, the system switches to the original tracking procedure, which refines the pose estimation using the last pose.

\subsection{Depth-Free Tracking Recovery}
When depth information is unavailable, the key differnce is that we can no longer apply Eq.~\eqref{eq:lost_tracking_condition} to verify whether the object has lost tracking. Instead, we must rely solely on RGB image data. In this case, tracking is considered lost if either of the following conditions is met:

\begin{itemize}
    \item The estimated Euclidean distance between the current and predicted object centers in the image plane exceeds a threshold:
    
    \begin{equation}
        \sqrt{(u-u_p)^2+(v-v_p)^2} > \theta_1.
    \end{equation}
    
    This condition ensures that the object's estimated position does not drift significantly from the predicted position. A large deviation suggests tracking failure.
    
    \item The difference in area between the observed object mask and the  rendered mask exceeds a threshold:
    
    \begin{equation}
        |Area(\text{mask}) - Area(\text{rendered\_mask})| > \theta_2.
    \end{equation}
    This condition accounts for changes in object size, which can occur due to incorrect scale estimation or occlusions. If the observed mask deviates significantly from the expected size, tracking is likely unreliable.
\end{itemize}

In case of tracking loss,  similar recovery strategies such as reinitialization, and hypothesis sampling can be applied to refine the object's pose and restore accurate tracking. \ours uses Kalman filter to roughly predict the object's depth (\(z\)) and rotation, then generate 20 pose hypotheses around the coarse estimation to improve accuracy.

\section{Dealing with CAD models of Incorrect Scale}\label{sec:incorrect_scale}
We now address the issue of using a CAD model with an incorrect scale. 
While recent state-of-the-art generative models like One-2-3-45++~\cite{liu2024one}  can generate mesh models from single or multiple-view images, these models are often produced at unknown scales, which makes them unsuitable for use in pose estimation tasks.

To recover the target object's metric scale, \ours assumes that the true depth is available for the initial frame. Otherwise, the method operates in a relative depth space obtained from any monocular depth estimation model. We employ a binary search procedure, similar to Alg.~\ref{alg:binary_search}, to determine the correct scale. By rendering and comparing images, we iteratively search for the optimal scaling factor that ensures the rendered image best matches the input.

\section{Experiments}\label{sec:eval}
We evaluated our proposed methods on an Intel\textsuperscript{\textregistered} Core\textsuperscript{TM} i9-10900K CPU at 3.7GHz with an RTX-3090 GPU.

\subsection{Metrics}
We follow the BOP Challenge 2019 protocol \cite{hodavn2020bop} to evaluate 6D object localization. The pose estimation error is measured using three pose-error functions:  

\begin{itemize}
    \item \textbf{Visible Surface Discrepancy (VSD)} – considers only the visible part to resolve pose ambiguity.
    \item \textbf{Maximum Symmetry-Aware Surface Distance (MSSD)} – accounts for global object symmetries and measures 3D surface deviation.
    \item \textbf{Maximum Symmetry-Aware Projection Distance (MSPD)} – incorporates object symmetries and evaluates 2D projection deviation.
\end{itemize}  

A pose is deemed correct under function \( e \) if \( e < \theta_e \), where \( e \in \{\text{VSD}, \text{MSSD}, \text{MSPD}\} \). The fraction of correctly estimated poses among all annotated instances defines the \textbf{Recall}.  

The \textbf{Average Recall} for a function \( e \), denoted \( AR_e \), is computed as the mean Recall over multiple thresholds \( \theta_e \), with VSD also considering a misalignment tolerance \( \tau \). Specifically:  

\begin{itemize}
    \item \( AR_{\text{VSD}} \) averages Recall over \( \tau \in [5\%, 50\%] \) (step 5\%) of the object diameter and \( \theta_{\text{VSD}} \in [0.05, 0.5] \) (step 0.05).  
    \item \( AR_{\text{MSSD}} \) averages Recall over \( \theta_{\text{MSSD}} \in [5\%, 50\%] \) (step 5\%) of the object diameter.  
    \item \( AR_{\text{MSPD}} \) averages Recall over \( \theta_{\text{MSPD}} \in [5r, 50r] \) (step \( 5r \)), where \( r = w/640 \) and \( w \) is the image width.  
\end{itemize}  
The overall method accuracy is given by:  

\begin{equation}
    AR = \frac{AR_{\text{VSD}} + AR_{\text{MSSD}} + AR_{\text{MSPD}}}{3}
\end{equation}  

\renewcommand{\arraystretch}{1.2} 
\begin{table*}[!htpb]
    \centering
    \fontsize{6.15}{8.2}\selectfont 
    \begin{tabular}{|p{2.3cm}!{\vrule width 2pt}ccc!{\vrule width 2pt}ccc!{\vrule width 2pt}ccc!{\vrule width 2pt}ccc|}
        \hline
        \textbf{Name} & \multicolumn{3}{c!{\vrule width 2pt}}{\textbf{FoundationPose (RGBD)}} & \multicolumn{3}{c!{\vrule width 2pt}}{\textbf{\ours (RGB)}} & \multicolumn{3}{c!{\vrule width 2pt}}{\textbf{Last Depth (RGB)}} & \multicolumn{3}{c|}{\textbf{FoundPose (RGB)}} \\ 
        \cline{2-13}
        & \textbf{Rec$(\uparrow)$} & \textbf{T.Err$(\downarrow)$} & \textbf{R.Err$(\downarrow)$} 
        & \textbf{Rec$(\uparrow)$} & \textbf{T.Err$(\downarrow)$} & \textbf{R.Err$(\downarrow)$}  & \textbf{Rec$(\uparrow)$} & \textbf{T.Err$(\downarrow)$} & \textbf{R.Err$(\downarrow)$}
      
         & \textbf{Rec$(\uparrow)$} & \textbf{T.Err$(\downarrow)$} & \textbf{R.Err$(\downarrow)$} \\ 
        \hline
        bleach\_hard\_00\_03 & 0.926 & 0.0072 & 3.68° & 0.856 & 0.016 & 4.22° & 0.015 & 0.490 & 149.63° & 0.0036 & 0.310 & 31.81° \\ \hline
        bleach0 & 0.956 & 0.0060 & 4.29° & 0.471 & 0.110 & 9.30° & 0.256 & 0.759 & 56.10° & 0.0052 & 0.369 & 21.84° \\ \hline
        mustard0 & 0.998 & 0.0031 & 3.20° & 0.875 & 0.020 & 3.64° & 0.099 & 0.693 & 13.89° & 0.0072 & 0.561 & 41.31° \\ \hline
        mustard\_easy & 0.979 & 0.0038 & 2.54° & 0.883 & 0.016 & 3.45° & 0.564 & 0.065 & 13.89° & 0.0083 & 0.474 & 51.51° \\ \hline
        sugar\_box1 & 0.973 & 0.0042 & 3.67° & 0.722 & 0.049 & 11.01° & 0.192 & 0.630 & 67.21° & 0.0083 & 0.317 & 31.61° \\ \hline
        sugar\_box\_yalehand0 & 0.331 & 0.0078 & 176.85° & 0.238 & 0.065 & 73.97° & 0.186 & 0.675 & 100.02° & 0.0079 & 0.328 & 41.18° \\ \hline
        cracker\_box\_reorient & 0.984 & 0.0038 & 2.73° & 0.861 & 0.022 & 3.34° & 0.444 & 0.549 & 8.14° & 0.0060 & 0.318 & 32.01° \\ \hline
        cracker\_box\_yalehand0 & 0.943 & 0.0101 & 3.20° & 0.791 & 0.074 & 6.30° & 0.135 & 0.823 & 73.02° & 0.0044 & 0.369 & 42.21° \\ \hline
        tomato\_soup\_can & 0.896 & 0.0062 & 8.53° & 0.714 & 0.025 & 9.06° & 0.066 & 0.743 & 65.39° & 0.0073 & 0.431 & 42.41° \\ \hline
    \end{tabular}
    \caption{Comparison of various methods based on average recall (Rec), translation error (T.Err), and rotation error (R.Err) for object pose estimation and tracking on the YCBInEAOT dataset ~\cite{wen2020se}.}
    \label{tab:ycbineaot}
\end{table*}
\renewcommand{\arraystretch}{1.0} 

In addition to recall scores, translation and rotation errors are also reported.
\subsection{Evaluation on YCBInEOAT Dataset}


In this section, we evaluate our proposed methods on the YCBInEOAT dataset~\cite{wen2020se}, which is a well-established benchmark for RGBD-based 6D pose tracking in robotic manipulation. We compare \ours with depth-based FoundationPose, one additional depth-free variant, and FoundPose~\cite{ornek2024foundpose}, a recent RGB-based foundation model.

Tab.~\ref{tab:ycbineaot} presents quantitative results, where we assess the methods based on recall (Rec$\uparrow$), translation error (T.Err$\downarrow$), and rotation error (R.Err$\downarrow$). The methods compared are:
\begin{itemize}
    \item \textbf{FoundationPose (RGBD)}: The original FoundationPose model using both RGB and depth as input.
    \item \textbf{\ours (RGB)}: Our enhanced depth-free method, where  depth input is replaced with a zero-depth matrix.
    \item \textbf{Last Depth (RGB)}: A variant that utilizes the previously estimated pose to render the last frame's depth, incorporating it as the depth input for the current frame. 
    \item \textbf{FoundPose (RGB)}: A recent foundation model that generates model templates in multiple views and extracts features using DINO-V2~\cite{oquab2023dinov2}, followed by KNN~\cite{guo2003knn} and PnP~\cite{moreno2007accurate} for pose estimation.
\end{itemize}

The results show that FoundationPose (RGBD) achieves the best overall performance, benefiting from depth input for precise pose estimation. However, our depth-free variant, \ours, demonstrates competitive performance, with recall values closely matching the RGBD-based approach in most cases. 
The Last Depth (RGB) method suffers from high translation and rotation errors, indicating that relying on the previous frame’s depth estimation is not a reliable alternative, especially due to error accumulation during the tracking process. FoundPose (RGB) struggles to achieve a high recall score, likely due to its reliance on template matching, which provides only a coarse pose estimate. An example is shown in Fig.~\ref{fig:tracking_example} and in the supplementary demo video.

Because the benchmark videos were taken under conditions where good depth information can be captured, it is no surprise that the FoundationPose baseline did better in these settings. On the other hand, when depth sensing is unreliable, e.g., as shown in Fig.~\ref{fig:example} and the accompanying video, \ours significantly outperforms depth-based method. 
Overall, \ours proves to be the best RGB-only solution, providing a promising alternative to depth sensor-dependent approaches. This makes it a top choice for scenarios where depth data is unavailable, unreliable, or too costly to obtain.

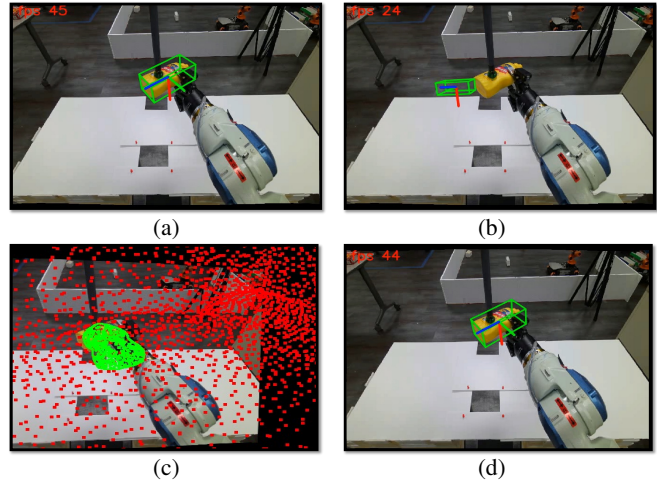
\begin{figure}[!htbp]
    \centering
    \begin{overpic}               
        [width=1\linewidth]{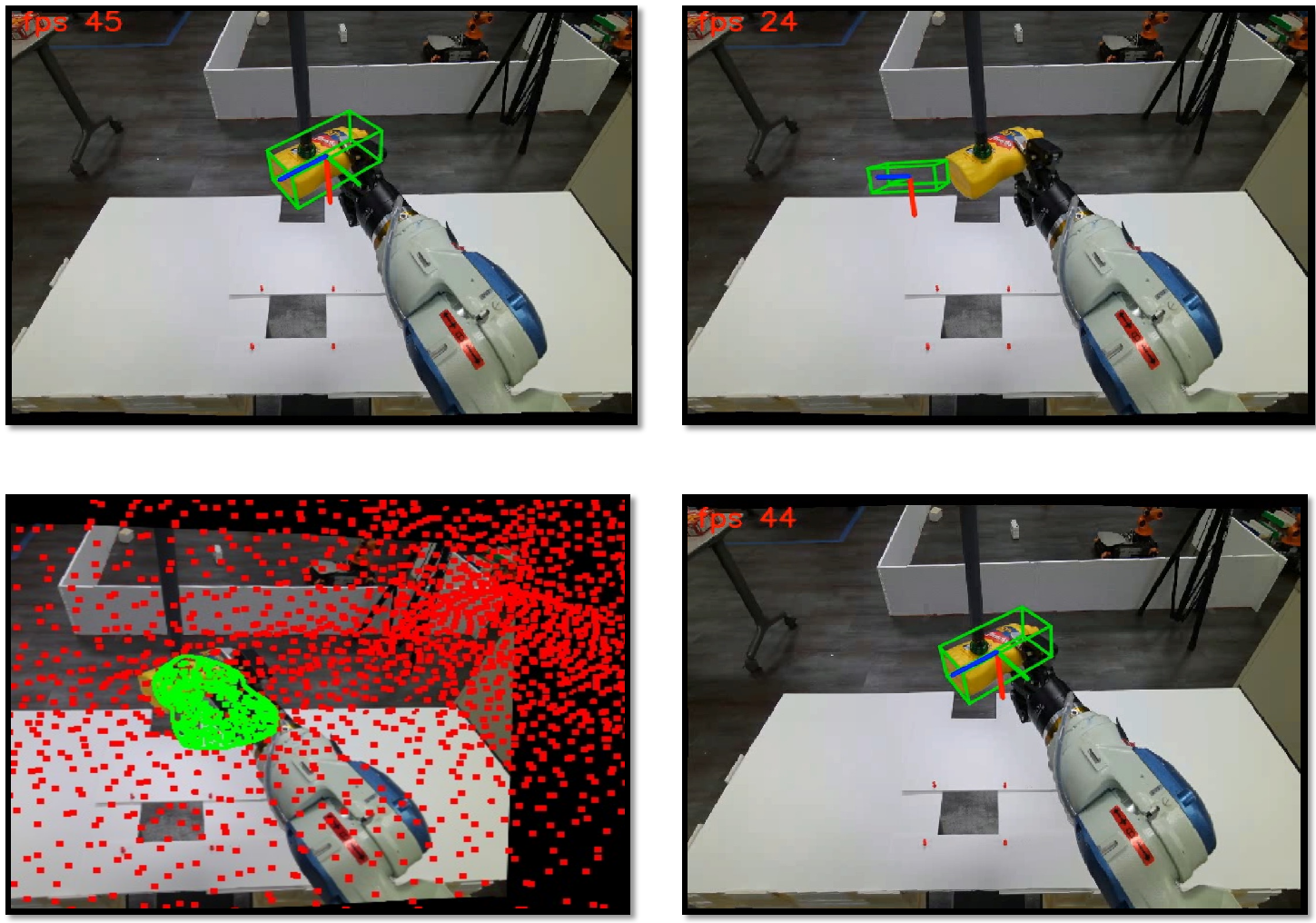}
             \small
             \put(22.5, 34.) {(a)}
             \put(72.5, 34.) {(b)}
             \put(22.5, -3) {(c)}
             \put(72.5, -3) {(d)}
        \end{overpic}
    \vspace{-2mm}
    \caption{(a) FoundationPose with RGBD input.
(b) FoundationPose leveraging the last estimated pose to render depth as input for the current frame. This variant is prone to lose tracking due to the accumulation of the drift error.
(c) FoundPose using only RGB input. The FoundPose is not very precise as the feature matching is not accurate enough.
(d) \ours which uses binary search for the initial pose registration and disables the depth channel by replacing it with a zero-depth matrix in the tracking. }
    \label{fig:tracking_example}
\end{figure}

\subsection{Tracking Loss Recovery}
\begin{figure}[t!]
    \centering
      \begin{overpic}               
        [width=1\linewidth]{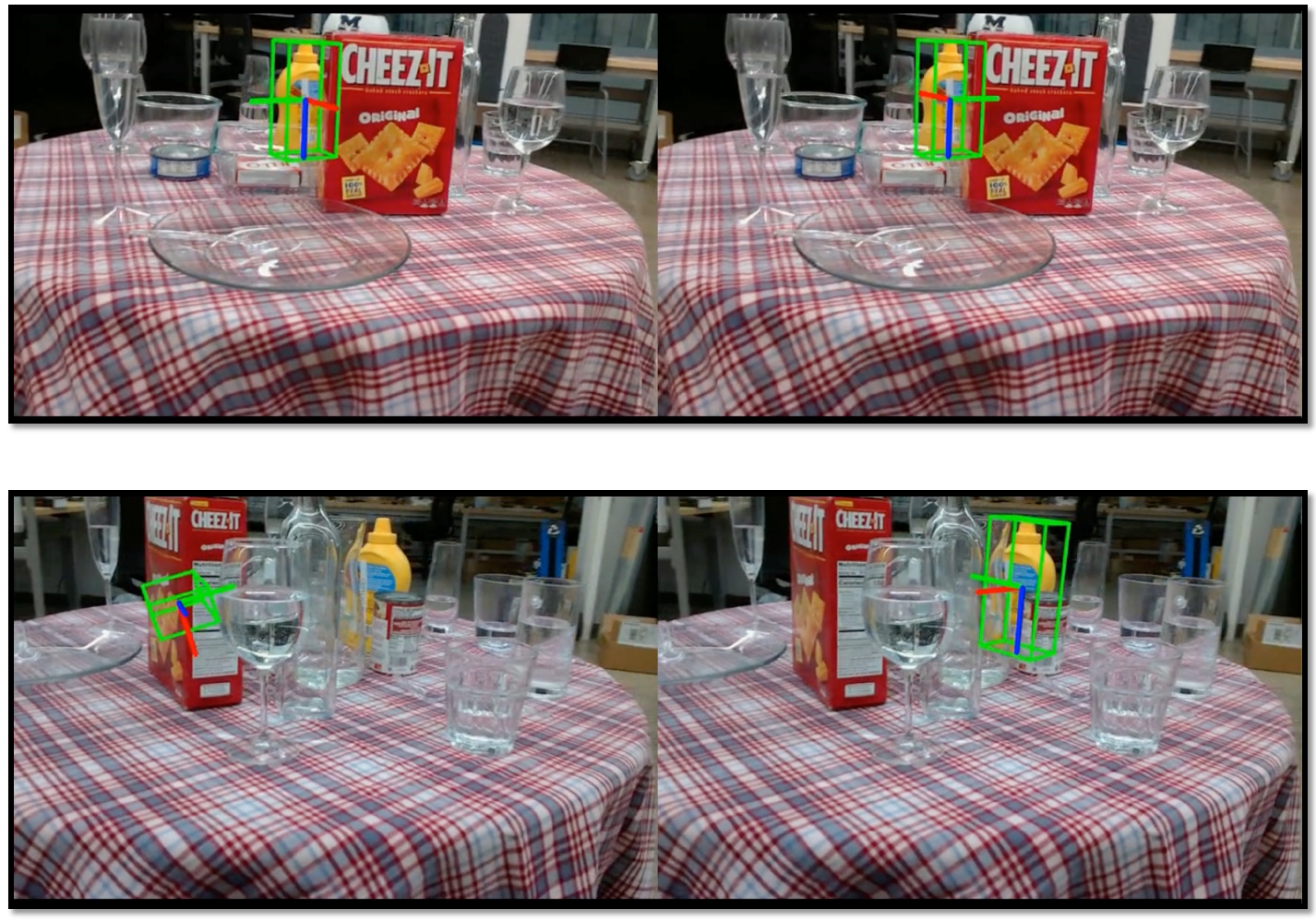}
             \small
             \put(22.5, 34) {(a)}
             \put(72.5, 34) {(b)}
             \put(22.5, -3) {(c)}
             \put(72.5, -3) {(d)}
        \end{overpic}
    \vspace{-1mm}
    \caption{Tracking recovery example: (a) and (c) show FoundationPose before and after tracking loss, while (b) and (d) illustrate that \ours's tracking recovery mechanism before and after object occlusion has occurred.}
    \label{fig:tracking_recover_example}
\end{figure}

\begin{figure}
    \centering
    \includegraphics[width=1\linewidth]{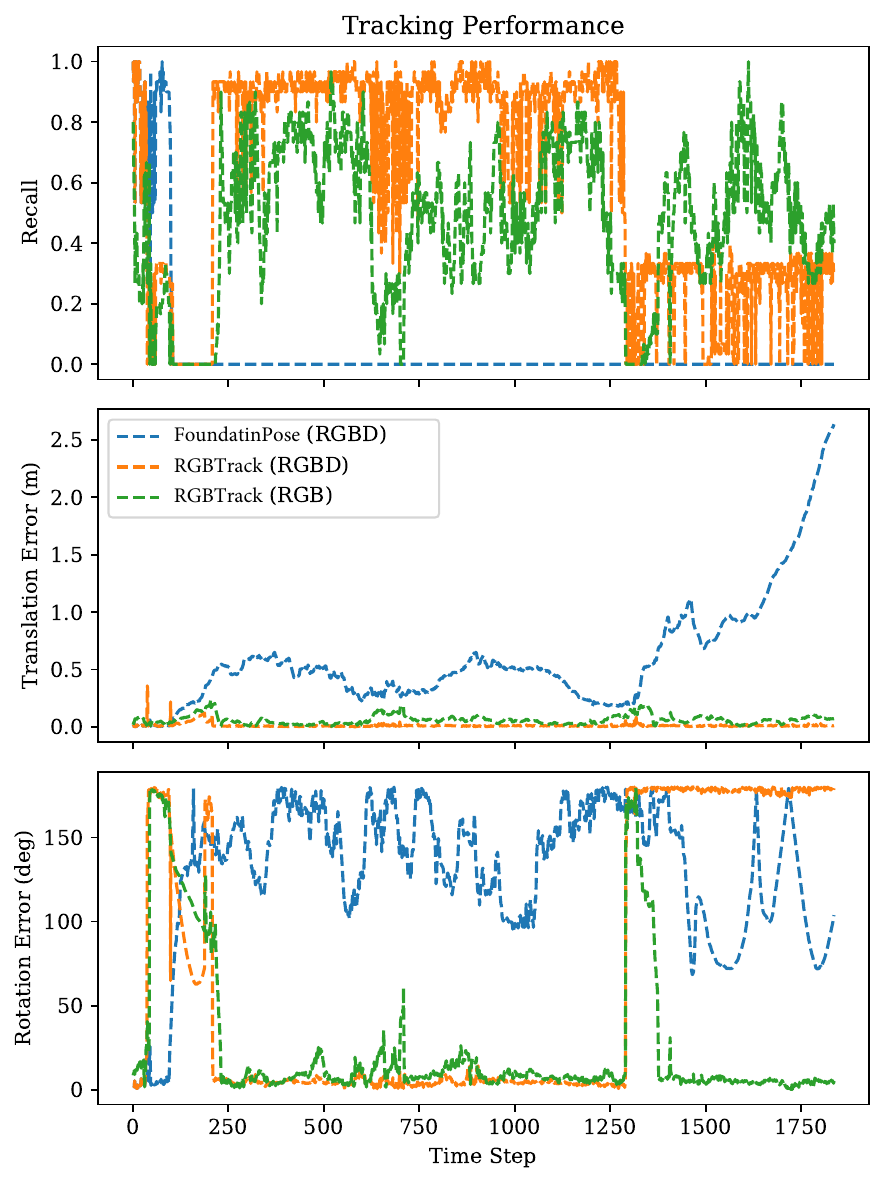}
    \vspace{-6mm}
    \caption{The Performance of Pose Tracking Methods with/without recovery mechanism for the mustard bottle object.}
    \label{fig:tracking_recover_plot}
\end{figure}

In this section, we evaluate the tracking recovery mechanism of \ours. We use XMem~\cite{cheng2022xmem} as our 2D object tracking module. However, other object tracking networks, such as SAM2~\cite{ravi2024sam} or YOLO~\cite{khanam2024yolov11}, can also be utilized. For this evaluation, we select scene 2 from set 8 in the ClearPose dataset~\cite{chen2022clearpose} because (1) CAD models of the opaque objects are available, (2) tracking methods are prone to failure due to occlusions from other objects, and (3) ground truth poses for all objects are provided. We note that there are a very limited number of such scenes in openly available datasets (a few more examples can be found in the accompanying video). Results are shown in Fig.\ref{fig:tracking_recover_example} and Fig.\ref{fig:tracking_recover_plot}. While the FoundationPose baseline loses tracking and fails to recover, our proposed framework successfully restores tracking.

The average performance across all frames, including those where tracking is completely lost, and all opaque objects is summarized in Tab.~\ref{tab:clearpose_average}. The FoundationPose baseline yields the lowest recall score, whereas variants equipped with the recovery mechanism demonstrate significantly higher recall. Among these, the depth-assisted variant outperforms its depth-free counterpart in terms of recall, exhibiting notably lower translation errors but higher rotation errors. \ours in RGB-only mode demonstrates a nicely balanced reduction of both translation and rotational errors. Despite incorporating XMem and a Kalman filter, our \ours framework maintains real-time tracking performance, running at approximately 22 FPS on average on an RTX 3090 at a $640\times 480$ image resolution while FoundationPose runs at $\sim45$ FPS.


\renewcommand{\arraystretch}{1.2} 
\begin{table}[t!]
    \centering
    \caption{Average performance comparison of different methods across all the frames in ClearPose set 8 scene 2.}
    \label{tab:clearpose_average}
    \renewcommand{\arraystretch}{1.1} 
    \setlength{\tabcolsep}{4pt} 
    \begin{tabular}{lccc}
        \toprule
        Metric & FoundationPose & \ours ($+$ depth) & \;\ours\;\;\\
        \midrule
        Recall & 0.0442 & 0.616 & 0.4682 \\
        T.Err (m) & 0.598 & 0.0129 & 0.0620 \\
        R.Err (°) & 130.36 & 67.795 & 25.806 \\
        \bottomrule
    \end{tabular}
\end{table}
\renewcommand{\arraystretch}{1.0} 

\subsection{Scale Recovery}
In this section, we evaluate the performance of the proposed scale recovery module on the YCBInEOAT dataset. During the evaluation, each object's CAD model is scaled by a factor of three before running \ours with and without the scale recovery module. The results, presented in Tab.~\ref{tab:unknown_scale}, demonstrate that the scale recovery module successfully restores the true scale of the model, especially for those scenarios where the object is not occluded in the first frame, allowing \ours to achieve accurate pose estimation. The recovery module fails on cracker\_box\_yalehand0,  tomato\_soup\_can\_yalehand0, and sugar\_box\_yalehand0 because in both scenarios the object is partially occluded by the robot hand. In contrast, when using an incorrectly scaled model, \ours (and FoundationPose as well) suffers and gets zero recall scores in all scenarios. This capability enables \ours to work with models generated by generative methods such as One-2-3-45++~\cite{liu2024one}, which often have unknown scales. A demonstration video is included in the supplementary material that best illustrates this feature.
\ifarxiv
\renewcommand{\arraystretch}{1.25} 
\begin{table}[t!]
    \centering
    \fontsize{6.5}{7}\selectfont 
    \setlength{\tabcolsep}{2pt} 
    \begin{tabular}{|p{2.5cm}!{\vrule width 2pt}ccc!{\vrule width 2pt}ccc|}
        \hline
        \textbf{Name} & \multicolumn{3}{c!{\vrule width 2pt}}{\textbf{With Scale Recovery}} & \multicolumn{3}{c|}{\textbf{Without}}  \\ 
        \cline{2-7}
        & \textbf{Rec$(\uparrow)$} & \textbf{T.Err$(\downarrow)$} & \textbf{R.Err$(\downarrow)$} 
        & \textbf{Rec$(\uparrow)$} & \textbf{T.Err$(\downarrow)$} & \textbf{R.Err$(\downarrow)$}  \\ 
        \hline
        bleach\_hard\_00\_03 & 0.929 & 0.0075 & 3.60°& 0.000 & 0.622 & 121°  \\ \hline
        bleach0 & 0.913 & 0.0091 & 5.73° & 0.000 & 0.353 & 145°  \\ \hline
        mustard0 & 0.985 & 0.0026 & 4.10° & 0.000 & 0.760 & 164°  \\ \hline
        mustard\_easy & 0.976 & 0.0042 & 3.23° & 0.000 & 0.397 & 161°  \\ \hline
        sugar\_box1 & 0.939 & 0.0056 & 4.01° & 0.000 & 0.615 & 158° \\ \hline
        sugar\_box\_yalehand0 & 0.000 & 0.118 & 91.47°& 0.000 & 0.494 & 120°  \\ \hline
        cracker\_box\_reorient & 0.989 & 0.0039 & 2.58° & 0.000 & 0.613 & 124° \\ \hline
        cracker\_box\_yalehand0 & 0.083 & 0.226 & 102.48° & 0.000 & 0.450 & 125° \\ \hline
        tomato\_soup\_can & 0.846 & 0.0084 & 10.46° & 0.000 & 0.281 & 138°  \\ \hline
    \end{tabular}
    \caption{Comparison of various methods based on average recall (Rec), translation error (T.Err), and rotation error (R.Err) with/without the scale recovery module.}
    \label{tab:unknown_scale}
\end{table}
\renewcommand{\arraystretch}{1.0} 
\fi

\renewcommand{\arraystretch}{1.25} 
\begin{table}[t!]
    \centering
    \fontsize{6}{6.5}\selectfont 
    \setlength{\tabcolsep}{2pt} 
    \begin{tabular}{|p{2.5cm}!{\vrule width 2pt}ccc!{\vrule width 2pt}ccc|}
        \hline
        \textbf{Name} & \multicolumn{3}{c!{\vrule width 2pt}}{\textbf{With Scale Recovery}} & \multicolumn{3}{c|}{\textbf{Without}}  \\ 
        \cline{2-7}
        & \textbf{Rec$(\uparrow)$} & \textbf{T.Err$(\downarrow)$} & \textbf{R.Err$(\downarrow)$} 
        & \textbf{Rec$(\uparrow)$} & \textbf{T.Err$(\downarrow)$} & \textbf{R.Err$(\downarrow)$}  \\ 
        \hline
        bleach\_hard\_00\_03 & 0.853 & 0.0169 & 4.18°& 0.000 & 0.622 & 121°  \\ \hline
        bleach0 & 0.433 & 0.117 & 9.286° & 0.000 & 0.353 & 145°  \\ \hline
        mustard0 & 0.726 & 0.0389 & 3.600° & 0.000 & 0.760 & 164°  \\ \hline
        mustard\_easy & 0.673 & 0.043 & 3.442° & 0.000 & 0.397 & 161°  \\ \hline
        sugar\_box1 & 0.530 & 0.087 & 8.34° & 0.000 & 0.615 & 158° \\ \hline
        sugar\_box\_yalehand0 & 0.000 & 1.018 & 116.47°& 0.000 & 0.494 & 120°  \\ \hline
        cracker\_box\_reorient & 0.894 & 0.0179 & 3.353° & 0.000 & 0.613 & 124° \\ \hline
        cracker\_box\_yalehand0 & 0.003 & 0.315 & 6.422° & 0.000 & 0.450 & 125° \\ \hline
        tomato\_soup\_can\_yalehand0 & 0.299 & 0.096 & 9.096° & 0.000 & 0.281 & 138°  \\ \hline
    \end{tabular}
    \caption{Comparison of various methods based on average recall (Rec), translation error (T.Err), and rotation error (R.Err) with/without the scale recovery module.}
    \label{tab:unknown_scale}
\end{table}
\renewcommand{\arraystretch}{1.0} 

\section{Conclusion}\label{sec:conclusion}



In this paper, we introduce, \ours, a state-of-the-art approach for real-time 6D pose estimation and tracking that relies \emph{exclusively on RGB input}, thereby eliminating the need for depth sensors. This depth-free strategy not only simplifies hardware requirements—making it ideal for mobile robotics/AR and cost- or power-constrained systems—but also offers robust performance in challenging environments where depth sensors may falter. Our method incorporates a carefully-designed and tuned recovery mechanism for tracking loss that employs a state machine combined with predictive filtering, ensuring that tracking can be promptly reinitialized even in dynamic scenes. Moreover, our approach dynamically adapts to CAD models of unknown scale by using an initial depth estimate from the first frame, which makes it particularly suited for real-world objects when precise dimensions are not available.

Despite these advantages, our method is not without limitations. As a depth-free method, \ours still has some gap when compared with FoundationPose when good depth sensing information is available, as shown in Tab.~\ref{tab:ycbineaot}. The depth estimation component, which aligns rendered and observed masks, can be less accurate when a large portion of the target object is occluded. Additionally, the recovery module, while effective, relies heavily on the accuracy of the 2D object tracking network, and its computational overhead can reduce the overall tracking frame rate.

Looking forward, we plan to extend \ours to operate in even more complex environments, refine our pose estimation algorithms to be fully on-par with depth-assisted methods, and improve generalization across a wider range of objects and conditions. Overall, we believe our approach holds strong potential for transformative applications in robotics, augmented reality, and broader computer vision domains.



\bibliographystyle{IEEEtran}
\bibliography{all}

\begin{thebibliography}{10}
\providecommand{\url}[1]{#1}
\csname url@samestyle\endcsname
\providecommand{\newblock}{\relax}
\providecommand{\bibinfo}[2]{#2}
\providecommand{\BIBentrySTDinterwordspacing}{\spaceskip=0pt\relax}
\providecommand{\BIBentryALTinterwordstretchfactor}{4}
\providecommand{\BIBentryALTinterwordspacing}{\spaceskip=\fontdimen2\font plus
\BIBentryALTinterwordstretchfactor\fontdimen3\font minus
  \fontdimen4\font\relax}
\providecommand{\BIBforeignlanguage}[2]{{%
\expandafter\ifx\csname l@#1\endcsname\relax
\typeout{** WARNING: IEEEtran.bst: No hyphenation pattern has been}%
\typeout{** loaded for the language `#1'. Using the pattern for}%
\typeout{** the default language instead.}%
\else
\language=\csname l@#1\endcsname
\fi
#2}}
\providecommand{\BIBdecl}{\relax}
\BIBdecl

\bibitem{wen2024foundationpose}
B.~Wen, W.~Yang, J.~Kautz, and S.~Birchfield, ``Foundationpose: Unified 6d pose
  estimation and tracking of novel objects,'' in \emph{Proceedings of the
  IEEE/CVF Conference on Computer Vision and Pattern Recognition}, 2024, pp.
  17\,868--17\,879.

\bibitem{liu2024one}
M.~Liu, C.~Xu, H.~Jin, L.~Chen, M.~Varma~T, Z.~Xu, and H.~Su, ``One-2-3-45: Any
  single image to 3d mesh in 45 seconds without per-shape optimization,''
  \emph{Advances in Neural Information Processing Systems}, vol.~36, 2024.

\bibitem{liu2023zero}
R.~Liu, R.~Wu, B.~Van~Hoorick, P.~Tokmakov, S.~Zakharov, and C.~Vondrick,
  ``Zero-1-to-3: Zero-shot one image to 3d object,'' in \emph{Proceedings of
  the IEEE/CVF international conference on computer vision}, 2023, pp.
  9298--9309.

\bibitem{liu2023one2345}
M.~Liu, C.~Xu, H.~Jin, L.~Chen, M.~V. T, Z.~Xu, and H.~Su, ``One-2-3-45: Any
  single image to 3d mesh in 45 seconds without per-shape optimization,''
  \emph{arXiv preprint arXiv:2306.16928}, 2023.

\bibitem{jun2023shap}
H.~Jun and A.~Nichol, ``Shap-e: Generating conditional 3d implicit functions,''
  \emph{arXiv preprint arXiv:2305.02463}, 2023.

\bibitem{cheng2022xmem}
H.~K. Cheng and A.~G. Schwing, ``Xmem: Long-term video object segmentation with
  an atkinson-shiffrin memory model,'' in \emph{European Conference on Computer
  Vision}.\hskip 1em plus 0.5em minus 0.4em\relax Springer, 2022, pp. 640--658.

\bibitem{peng2019pvnet}
S.~Peng, Y.~Liu, Q.~Huang, X.~Zhou, and H.~Bao, ``Pvnet: Pixel-wise voting
  network for 6dof pose estimation,'' in \emph{Proceedings of the IEEE/CVF
  conference on computer vision and pattern recognition}, 2019, pp. 4561--4570.

\bibitem{hu2020single}
Y.~Hu, P.~Fua, W.~Wang, and M.~Salzmann, ``Single-stage 6d object pose
  estimation,'' in \emph{Proceedings of the IEEE/CVF conference on computer
  vision and pattern recognition}, 2020, pp. 2930--2939.

\bibitem{hu2019segmentation}
Y.~Hu, J.~Hugonot, P.~Fua, and M.~Salzmann, ``Segmentation-driven 6d object
  pose estimation,'' in \emph{Proceedings of the IEEE/CVF conference on
  computer vision and pattern recognition}, 2019, pp. 3385--3394.

\bibitem{li2019cdpn}
Z.~Li, G.~Wang, and X.~Ji, ``Cdpn: Coordinates-based disentangled pose network
  for real-time rgb-based 6-dof object pose estimation,'' in \emph{Proceedings
  of the IEEE/CVF international conference on computer vision}, 2019, pp.
  7678--7687.

\bibitem{rad2017bb8}
M.~Rad and V.~Lepetit, ``Bb8: A scalable, accurate, robust to partial occlusion
  method for predicting the 3d poses of challenging objects without using
  depth,'' in \emph{Proceedings of the IEEE international conference on
  computer vision}, 2017, pp. 3828--3836.

\bibitem{tekin2018real}
B.~Tekin, S.~N. Sinha, and P.~Fua, ``Real-time seamless single shot 6d object
  pose prediction,'' in \emph{Proceedings of the IEEE conference on computer
  vision and pattern recognition}, 2018, pp. 292--301.

\bibitem{zakharov2019dpod}
S.~Zakharov, I.~Shugurov, and S.~Ilic, ``Dpod: 6d pose object detector and
  refiner,'' in \emph{Proceedings of the IEEE/CVF international conference on
  computer vision}, 2019, pp. 1941--1950.

\bibitem{wen2020se}
B.~Wen, C.~Mitash, B.~Ren, and K.~E. Bekris, ``se (3)-tracknet: Data-driven 6d
  pose tracking by calibrating image residuals in synthetic domains,'' in
  \emph{2020 IEEE/RSJ International Conference on Intelligent Robots and
  Systems (IROS)}.\hskip 1em plus 0.5em minus 0.4em\relax IEEE, 2020, pp.
  10\,367--10\,373.

\bibitem{wen2021bundletrack}
B.~Wen and K.~Bekris, ``Bundletrack: 6d pose tracking for novel objects without
  instance or category-level 3d models,'' in \emph{2021 IEEE/RSJ International
  Conference on Intelligent Robots and Systems (IROS)}.\hskip 1em plus 0.5em
  minus 0.4em\relax IEEE, 2021, pp. 8067--8074.

\bibitem{sun2022onepose}
J.~Sun, Z.~Wang, S.~Zhang, X.~He, H.~Zhao, G.~Zhang, and X.~Zhou, ``Onepose:
  One-shot object pose estimation without cad models,'' in \emph{Proceedings of
  the IEEE/CVF Conference on Computer Vision and Pattern Recognition}, 2022,
  pp. 6825--6834.

\bibitem{he2022onepose++}
X.~He, J.~Sun, Y.~Wang, D.~Huang, H.~Bao, and X.~Zhou, ``Onepose++:
  Keypoint-free one-shot object pose estimation without cad models,''
  \emph{Advances in Neural Information Processing Systems}, vol.~35, pp.
  35\,103--35\,115, 2022.

\bibitem{moreno2007accurate}
F.~Moreno-Noguer, V.~Lepetit, and P.~Fua, ``Accurate non-iterative o(n)
  solution to the pnp problem,'' in \emph{2007 IEEE 11th International
  Conference on Computer Vision}.\hskip 1em plus 0.5em minus 0.4em\relax Ieee,
  2007, pp. 1--8.

\bibitem{besl1992method}
P.~J. Besl and N.~D. McKay, ``A method for registration of 3-d shapes,''
  \emph{IEEE Transactions on pattern analysis and machine intelligence},
  vol.~14, no.~2, pp. 239--256, 1992.

\bibitem{drost2010model}
B.~Drost, M.~Ulrich, N.~Navab, and S.~Ilic, ``Model globally, match locally:
  Efficient and robust 3d object recognition,'' in \emph{2010 IEEE computer
  society conference on computer vision and pattern recognition}.\hskip 1em
  plus 0.5em minus 0.4em\relax Ieee, 2010, pp. 998--1005.

\bibitem{li2020cdpn}
Y.~Li, G.~Wang, X.~Ji, Y.~Xiang, and D.~Fox, ``Cdpn: Coordinates-based
  disentangled pose network for real-time rgb-based 6-dof object pose
  estimation,'' in \emph{Proceedings of the IEEE International Conference on
  Computer Vision (ICCV)}.\hskip 1em plus 0.5em minus 0.4em\relax IEEE, 2020,
  pp. 7678--7687.

\bibitem{ausserlechner2024zs6d}
P.~Ausserlechner, D.~Haberger, S.~Thalhammer, J.-B. Weibel, and M.~Vincze,
  ``Zs6d: Zero-shot 6d object pose estimation using vision transformers,'' in
  \emph{2024 IEEE International Conference on Robotics and Automation
  (ICRA)}.\hskip 1em plus 0.5em minus 0.4em\relax IEEE, 2024, pp. 463--469.

\bibitem{labbe2022megapose}
Y.~Labb{\'e}, L.~Manuelli, A.~Mousavian, S.~Tyree, S.~Birchfield, J.~Tremblay,
  J.~Carpentier, M.~Aubry, D.~Fox, and J.~Sivic, ``Megapose: 6d pose estimation
  of novel objects via render \& compare,'' \emph{arXiv preprint
  arXiv:2212.06870}, 2022.

\bibitem{cai2022ove6d}
D.~Cai, J.~Heikkil{\"a}, and E.~Rahtu, ``Ove6d: Object viewpoint encoding for
  depth-based 6d object pose estimation,'' in \emph{Proceedings of the IEEE/CVF
  Conference on Computer Vision and Pattern Recognition}, 2022, pp. 6803--6813.

\bibitem{lin2024sam}
J.~Lin, L.~Liu, D.~Lu, and K.~Jia, ``Sam-6d: Segment anything model meets
  zero-shot 6d object pose estimation,'' in \emph{Proceedings of the IEEE/CVF
  Conference on Computer Vision and Pattern Recognition}, 2024, pp.
  27\,906--27\,916.

\bibitem{nguyen2022templates}
V.~N. Nguyen, Y.~Hu, Y.~Xiao, M.~Salzmann, and V.~Lepetit, ``Templates for 3d
  object pose estimation revisited: Generalization to new objects and
  robustness to occlusions,'' in \emph{Proceedings of the IEEE/CVF conference
  on computer vision and pattern recognition}, 2022, pp. 6771--6780.

\bibitem{nguyen2024gigapose}
V.~N. Nguyen, T.~Groueix, M.~Salzmann, and V.~Lepetit, ``Gigapose: Fast and
  robust novel object pose estimation via one correspondence,'' in
  \emph{Proceedings of the IEEE/CVF Conference on Computer Vision and Pattern
  Recognition}, 2024, pp. 9903--9913.

\bibitem{okorn2021zephyr}
B.~Okorn, Q.~Gu, M.~Hebert, and D.~Held, ``Zephyr: Zero-shot pose hypothesis
  rating,'' in \emph{2021 IEEE International Conference on Robotics and
  Automation (ICRA)}.\hskip 1em plus 0.5em minus 0.4em\relax IEEE, 2021, pp.
  14\,141--14\,148.

\bibitem{ornek2024foundpose}
E.~P. {\"O}rnek, Y.~Labb{\'e}, B.~Tekin, L.~Ma, C.~Keskin, C.~Forster, and
  T.~Hodan, ``Foundpose: Unseen object pose estimation with foundation
  features,'' in \emph{European Conference on Computer Vision}.\hskip 1em plus
  0.5em minus 0.4em\relax Springer, 2024, pp. 163--182.

\bibitem{xiang2017posecnn}
Y.~Xiang, T.~Schmidt, V.~Narayanan, and D.~Fox, ``Posecnn: A convolutional
  neural network for 6d object pose estimation in cluttered scenes,''
  \emph{arXiv preprint arXiv:1711.00199}, 2017.

\bibitem{chen2022clearpose}
X.~Chen, H.~Zhang, Z.~Yu, A.~Opipari, and O.~Chadwicke~Jenkins, ``Clearpose:
  Large-scale transparent object dataset and benchmark,'' in \emph{European
  conference on computer vision}.\hskip 1em plus 0.5em minus 0.4em\relax
  Springer, 2022, pp. 381--396.

\bibitem{wen2023bundlesdf}
B.~Wen, J.~Tremblay, V.~Blukis, S.~Tyree, T.~M{\"u}ller, A.~Evans, D.~Fox,
  J.~Kautz, and S.~Birchfield, ``Bundlesdf: Neural 6-dof tracking and 3d
  reconstruction of unknown objects,'' in \emph{Proceedings of the IEEE/CVF
  Conference on Computer Vision and Pattern Recognition}, 2023, pp. 606--617.

\bibitem{bhat2023zoedepth}
S.~F. Bhat, R.~Birkl, D.~Wofk, P.~Wonka, and M.~M{\"u}ller, ``Zoedepth:
  Zero-shot transfer by combining relative and metric depth,'' \emph{arXiv
  preprint arXiv:2302.12288}, 2023.

\bibitem{yin2023metric3d}
W.~Yin, C.~Zhang, H.~Chen, Z.~Cai, G.~Yu, K.~Wang, X.~Chen, and C.~Shen,
  ``Metric3d: Towards zero-shot metric 3d prediction from a single image,'' in
  \emph{Proceedings of the IEEE/CVF International Conference on Computer
  Vision}, 2023, pp. 9043--9053.

\bibitem{ravi2024sam}
N.~Ravi, V.~Gabeur, Y.-T. Hu, R.~Hu, C.~Ryali, T.~Ma, H.~Khedr, R.~R{\"a}dle,
  C.~Rolland, L.~Gustafson \emph{et~al.}, ``Sam 2: Segment anything in images
  and videos,'' \emph{arXiv preprint arXiv:2408.00714}, 2024.

\bibitem{hodavn2020bop}
T.~Hoda{\v{n}}, M.~Sundermeyer, B.~Drost, Y.~Labb{\'e}, E.~Brachmann,
  F.~Michel, C.~Rother, and J.~Matas, ``Bop challenge 2020 on 6d object
  localization,'' in \emph{Computer Vision--ECCV 2020 Workshops: Glasgow, UK,
  August 23--28, 2020, Proceedings, Part II 16}.\hskip 1em plus 0.5em minus
  0.4em\relax Springer, 2020, pp. 577--594.

\bibitem{oquab2023dinov2}
M.~Oquab, T.~Darcet, T.~Moutakanni, H.~Vo, M.~Szafraniec, V.~Khalidov,
  P.~Fernandez, D.~Haziza, F.~Massa, A.~El-Nouby \emph{et~al.}, ``Dinov2:
  Learning robust visual features without supervision,'' \emph{arXiv preprint
  arXiv:2304.07193}, 2023.

\bibitem{guo2003knn}
G.~Guo, H.~Wang, D.~Bell, Y.~Bi, and K.~Greer, ``Knn model-based approach in
  classification,'' in \emph{On The Move to Meaningful Internet Systems 2003:
  CoopIS, DOA, and ODBASE: OTM Confederated International Conferences, CoopIS,
  DOA, and ODBASE 2003, Catania, Sicily, Italy, November 3-7, 2003.
  Proceedings}.\hskip 1em plus 0.5em minus 0.4em\relax Springer, 2003, pp.
  986--996.

\bibitem{khanam2024yolov11}
R.~Khanam and M.~Hussain, ``Yolov11: An overview of the key architectural
  enhancements,'' \emph{arXiv preprint arXiv:2410.17725}, 2024.

\end{thebibliography}

\end{document}